\pdfoutput=1

\documentclass[11pt]{article}

\usepackage[]{EACL2023}

\usepackage{times}
\usepackage{latexsym}
\usepackage{graphicx}
\usepackage{booktabs}
\usepackage{amsmath}
\usepackage{amssymb}
\usepackage{algorithmic}
\usepackage{algorithm}
\usepackage{cleveref}
\crefformat{section}{\S#2#1#3} 
\crefformat{subsection}{\S#2#1#3}
\crefformat{subsubsection}{\S#2#1#3}

\usepackage[T1]{fontenc}

\usepackage[utf8]{inputenc}

\usepackage{microtype}

\usepackage{inconsolata}
\usepackage{listings}
\usepackage{xcolor}

\definecolor{codegreen}{rgb}{0,0.6,0}
\definecolor{codegray}{rgb}{0.5,0.5,0.5}
\definecolor{codepurple}{rgb}{0.58,0,0.82}
\definecolor{backcolour}{rgb}{0.95,0.95,0.92}
\lstdefinestyle{mystyle}{
    commentstyle=\color{codegreen},
    keywordstyle=\color{magenta},
    numberstyle=\tiny\color{codegray},
    stringstyle=\color{codepurple},
    basicstyle=\ttfamily\footnotesize,
    breakatwhitespace=false,         
    breaklines=true,                 
    captionpos=b,                    
    keepspaces=true,                 
    numbers=left,                    
    numbersep=2pt,                  
    showspaces=false,                
    showstringspaces=false,
    showtabs=false,                  
    tabsize=4
}

\title{Multilingual Representation Distillation with Contrastive Learning}

\author{
    Weiting Tan$^{\spadesuit}$\thanks{\ \ Work was done during an internship at Meta AI Research.}  
\quad
\textbf{Kevin Heffernan}$^{\heartsuit}$
\quad
\textbf{Holger Schwenk}$^{\heartsuit}$
\quad 
\textbf{Philipp Koehn}$^{\spadesuit\heartsuit}$
\\
\\
$^{\spadesuit}$Johns Hopkins University \\
$^{\heartsuit}$Meta AI Research
    \\
    {\tt \{wtan12, phi\}@jhu.edu \{kevinheffernan,schwenk\}@meta.com}
}

\begin{document}
\maketitle
\begin{abstract}
Multilingual sentence representations from large models encode semantic information from two or more languages and can be used for different cross-lingual information retrieval and matching tasks. In this paper, we integrate contrastive learning into multilingual representation distillation and use it for quality estimation of parallel sentences (i.e., find semantically similar sentences that can be used as translations of each other). We validate our approach with multilingual similarity search and corpus filtering tasks. Experiments across different low-resource languages show that our method greatly outperforms previous sentence encoders such as LASER, LASER3, and LaBSE.
\end{abstract}

\section{Introduction}
\begin{figure}[t]
    \centering
    \begin{tabular}{c}
        \includegraphics[width=0.5\textwidth]{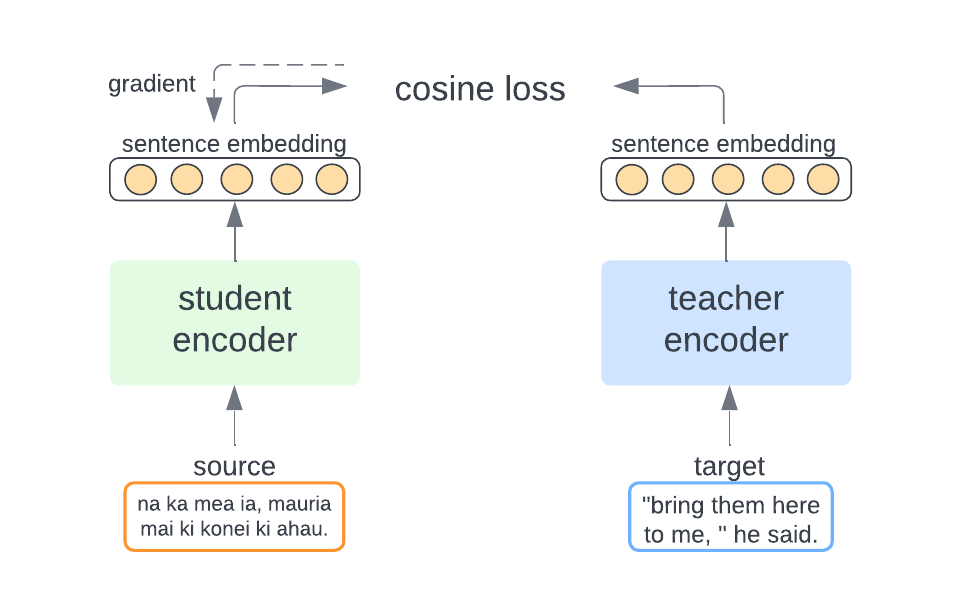} \\
    \end{tabular}
    \caption{Student-Teacher Distillation in LASER3.
    \label{fig:method-laser3}}
\end{figure}

With the rise of neural networks, high-dimensional word-level sequence representations play an important role in almost any natural language processing task. Contextual representations from large pre-trained language models \cite{transformer, bert, roberta, xlm} have shown advantages compared to earlier static embeddings \cite{word2vec, glove}. However, they are not pre-trained with sentence-level objectives and their representations of two different sentences cannot be easily used to indicate semantic similarity. To encode sentence-level information, LASER \cite{laser} pools a sentence embedding from the encoder and feeds it to the decoder. Another approach is to use siamese-structured models where two identical encoders are used to represent sentences of similar meaning and are trained to push their representations close to each other. Sentence-Transformers \cite{sbert} are siamese-structured models that are initialized with pre-trained large models like BERT \cite{bert} or Roberta \cite{roberta}. After fine-tuning, Sentence-Transformers improve their sentence representation for cross-lingual tasks. Besides fine-tuning with identical (siamese) encoders, distillation can be used to retrieve better representations. \citet{reimers} extends a monolingual sentence representation into a multilingual representation with model distillation. Similarly, \citet{laser3} proposed LASER3, a student-teacher architecture that distills information from a pre-trained teacher encoder to a student encoder in new languages. As shown in Figure \ref{fig:method-laser3}, the distillation process updates the student encoder with the gradient from the cosine loss. Note that it freezes the parameters of the teacher encoder, which is already pre-trained on the target language (English in our case). Therefore, the target sentence embedding is fixed and the corresponding source embedding is aligned with the target embedding.

\begin{figure*}[h]
    \centering
    \begin{tabular}{c}        \includegraphics[width=1.02\textwidth,trim={1cm 0 0 0}]{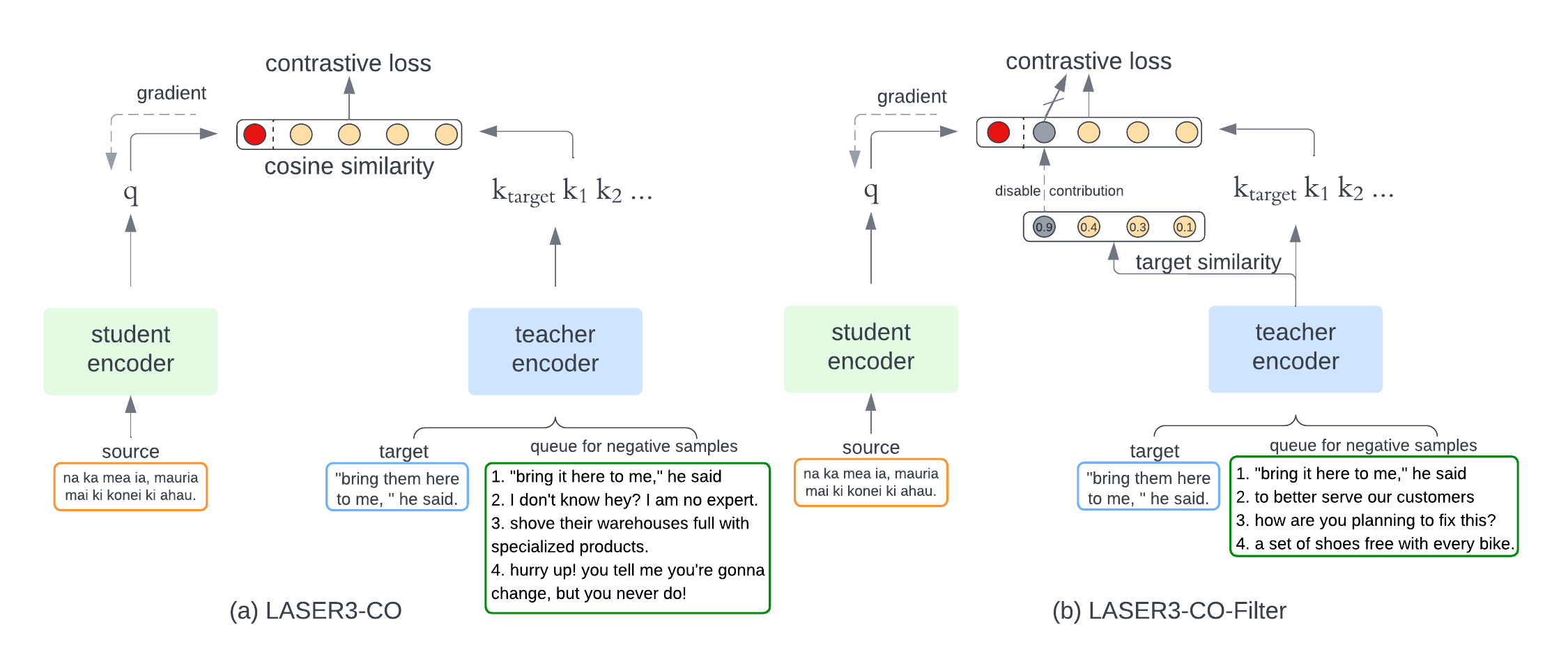} \\
    \end{tabular}
    \caption{Visual Explanation of LASER3-CO (vanilla) and LASER3-CO-Filter (filtered). Source and target come from input bitexts dataset and queue is constructed with earlier batch's samples. In the (b) filtered version, a pre-filtering mechanism is employed to filter out extremely negative samples from the queue.
    \label{fig:method}}
\end{figure*}
In this paper, we focus on the quality estimation of parallel sentences in low-resource languages, which requires models to distinguish similar and dissimilar sentence pairs. Contrastive learning is helpful because its objective not only aligns similar sentences' representations but also pushes away representations from dissimilar sentences, which enables the model to be more confident in similar sentences. Additionally, contrastive learning is a form of self-supervision that fits well in our low-resource setting where only a limited amount of clean data is available.

In practice, we integrate contrastive learning into the distillation-based architecture from LASER3 and use our contrastive distillation method to train encoders for low-resource languages. Inspired by \citet{moco} and \citet{memorycon}, we used a queue to store negative samples as self-supervision to train better encoders. We also employed a pre-filtering mechanism to find hard negative samples and showed that they benefit the distillation of sentence representations.

To evaluate different encoders, we rely on multilingual similarity search and corpus filtering tasks. In the multilingual similarity search task, we encode all source and target sentences and use a cosine-based similarity metric called margin score \cite{marginscore} to pair source and target sentences. In the corpus filtering task, a mined noisy parallel corpus is given and we use the encoder with margin score to compute a similarity score for each sentence pair. Then we filter out the low-score pairs and use the remaining corpus to train a neural machine translation model whose performance is evaluated with BLEU \cite{papineni-etal-2002-bleu}. Compared to previous works, we observe consistent improvement when using contrastive distillation. We also compared our approach with another simple but effective data augmentation technique, back-translation, and show that contrastive distillation achieves tied or better performance with less data.

\section{Approach}\label{sec::approach}
To train high-quality sentence representation, we integrate contrastive learning into sentence representation distillation and our contrastive distillation method is visualized in Figure \ref{fig:method}. The motivation for using contrastive learning is two-fold:
\begin{itemize}
    \item The self-supervision from contrastive learning helps representation learning in low-resource settings.
    \item Contrastive learning enables models to recognize similar and dissimilar sentences, which is crucial for filtering out noisy sentence pairs.
\end{itemize}

\subsection{LASER3-CO}\label{sec::moco}
This architecture corresponds to Figure \ref{fig:method} (a). We name it LASER3-CO since it integrates \underline{co}ntrastive learning to LASER3's distillation pipeline. We are inspired by \citet{moco} to use a queue to store contrastive (negative) samples. The negative samples are used to train encoders so that sentences of different meaning have dissimilar representations. During training, we use previous batches' target language sentences as negative samples and when there are too many negative samples in the queue, we remove the samples from the earliest batches. Our LASER3-CO approach has the following steps:
\begin{itemize}
    \item Pre-train LASER to use as the teacher encoder $\theta_t$ on high resource languages such as English. Randomly initialize the student encoder and perform distillation with the teacher encoder. After distillation, we obtain the pre-trained student encoder $\theta_s$.
    \item Fine-tune $\theta_s$ using queue(memory)-based\footnote{We use queue size N=4096 throughout experiments.} contrastive learning. For each input source sentence $x$ and target sentence $y$, encode them with student and teacher encoders respectively and we have their representations $q=\theta_s(x), k_{\text{target}} = k_{+}= \theta_t(y)$ (we use "+" as an abbreviation for the positive target sentence). We also encode all (N) negative sentences in the queue and have their representation $k_i = \theta_t(\text{queue}_i), i \in [1,N]$.
    \item Perform normalization on $q, k_{i}, i\in [0,N]$ and then compute the contrastive loss using infoNCE \cite{infonce}
    \begin{equation}
        \mathcal{L} = -\log \frac{\text{exp}(q\cdot k_{+} /\tau)}{\sum_{i=1}^{N} \text{exp}(q\cdot k_i /\tau)}
    \end{equation}
    Here $\tau$ is the temperature parameter\footnote{Empirically we found $\tau=0.05$ (widely used in previous literature) works well.} that controls the strength of the penalty. 
    \item Update student encoder $\theta_s$ with loss. Enqueue the most recent target language (English) sentences and dequeue the earliest sentences if the queue size exceeds the limit (N).
\end{itemize}

\noindent In LASER3-CO, our training process is very similar to MOCO \cite{moco}, though we do not use the momentum update for the teacher. Instead, we freeze the pre-trained teacher (LASER in our case) that already has a high-quality representation of English (or any other pivot language), so that English sentences are encoded consistently during distillation. Then we try to align the representation of the student encoder to the teacher, only allowing gradients to flow through the student encoder.

\subsection{LASER3-CO-Filter}\label{sec::laser3-cf}

This architecture corresponds to Figure \ref{fig:method} (b). The motivation is that previous work \cite{hardnegative} has shown hard negatives improve representation. In our task, for each parallel sentence pair $(x,y)$, the hard negatives would be sentences $y'$ that are hard to distinguish from the true target $y$ (we can also find hard negatives $x'$ in the source language, but because our teacher encoder has better representation for the target language, English, we decide to focus on target-side hard negatives only). Hard negatives are beneficial because they force the model to learn more complex features to distinguish them from the true target sentence.

To find more hard-negatives for contrastive fine-tuning, we change LASER3-CO in two ways: (1)~disable shuffling for the data loader and (2)~use a pre-filtering mechanism to filter out bad samples from the queue (we name this model LASER3-CO-Filter where Filter refers to the pre-filtering mechanism). Bad samples are hard negative samples $y'$ that are too similar to $y$ (e.g. "What is your name" versus "What's your name"). Treating these extremely similar sentences as negative samples would hurt the encoder's representation and we devise a pre-filtering method to filter them out.

\paragraph{Disable Shuffling} We sort sentences by length\footnote{Our implementation is based on Fairseq \cite{fairseq} which by default sorts the sentences by length} and disable the shuffling of data loader so that consecutive batches contain sentences of similar length. As our queue is updated by enqueuing the most recent batch and dequeuing the earliest batch, disabling shuffling makes the queue store sentences of similar length. Because all samples in the queue are used as negative samples to be contrasted with the true target sentence and because they are of a similar length, it is much more likely that we find hard negatives (quantitative analysis provided in \Cref{sec::analysis} and Figure \ref{fig:analysis-sim}).

\paragraph{Pre-filtering Mechanism} After disabling shuffling, we show that more hard-negatives are found. However, we also found that there are many cases of extremely hard negatives (e.g. "What is your name" versus "What's your name", "Bring them here to me" versus "Bring it here to me"). Though hard negatives help contrastive fine-tuning to obtain better sentence representation, extremely hard negatives would confuse the model and incorrectly update the parameters. Therefore we employ a simple pre-filtering mechanism to prevent these extremely hard negatives from contributing to the contrastive loss: After we encode the target sentence and sentences in the queue, we have the representations $k_{+}, k_{1},\cdots k_{N}$ as described in \Cref{sec::moco}. Then we compute the cosine similarity between the true target and sentences in the queue and get similarity scores $ \cos(k_{+}, k_i), i\in [1, N]$. For any negative sample i, if $\cos(k_{+}, k_i) \geq \sigma$ (where $\sigma$ is a pre-defined threshold hyper-parameter),\footnote{We use $\sigma=0.9$ throughout our experiments. Effects of different $\sigma$ values are analyzed in Figure \ref{fig:analysis-sim} (right)} we do not use it in the InfoNCE loss. In other words, let $S = \{i: \cos(k_{+}, k_i) < \sigma\}$ be a set that contains no extremely hard negatives for the threshold $\sigma$, the InfoNCE loss would be computed as

\begin{equation}
    \mathcal{L} = -\log \frac{\text{exp}(q\cdot k_{+} /\tau)}{\sum_{i \in S}\text{exp}(q\cdot k_i /\tau)}
\end{equation}
In practice, we perform this filtering step with batches of samples, as detailed in \cref{sec::pseudocode}. PyTorch-like pseudocode of the LASER3-CO-Filter model is shown below.
\begin{algorithm}
\caption{Pseudocode of LASER3-CO-Filter}
\label{algo:filter}
\begin{lstlisting}[language=Python, numbers=none]
# N: queue size;   bz: batch size
# S: size of set without extremely hard negatives
# t: temperature;   h: filtering threshold
# emb_dim: embedding dimension
# einsum: einsum function available in PyTorch
# queue: stores N earlier target embed
# teacher: pre-trained LASER encoder
# student: LASER3 student encoder
# filter: function that filters out extremely hard negatives (details available in appendix A)

# freeze teacher encoder
teacher.params.requires_grad = False 
for (source, target) in loader:
    src_emb = normalize(student(source),dim=1)
    tgt_emb = normalize(teacher(target),dim=1)
    neg_emb = normalize(teacher(queue),dim=1)
    #--------- Begin Pre-filtering ---------
    # bz x N (mask for indices to be kept)
    mask = (tgt_emb @ neg_emb.T < h).long()
    # bz x S x emb_dim
    neg_emb = filter(neg_emb, mask)
    #--------- End Pre-filtering ------------
    #postive logits: bz x 1
    l_pos = src_emb @ tgt_emb.T
    # negative logits: bz x S
    l_neg = einsum("bh,bsh->bs", src_emb, neg_emb)
    logits = cat([l_pos, l_neg],dim=1)
    loss = CrossEntropyLoss(logits/t, zeros(bz))
    loss.backward()
    update(student)
    enqueue(queue, tgt_emb)
    if queue.size > N:
        dequeue(queue)
\end{lstlisting}
\end{algorithm}

\section{Experiment on Low-resource Languages}\label{sec::low-resource}
\subsection{Dataset}\label{sec::dataset}
For training, we use two sources of data: existing clean data and back-translation data.

\paragraph{Clean data} Our clean data comes from publicly available datasets\footnote{All available online, mostly taken from OPUS: \url{https://opus.nlpl.eu/}} such as \textit{jw300}, \textit{bible} \cite{bible}, \textit{tatoeba} \cite{opus}, \textit{wikimedia} \cite{opus}, \textit{gv} \cite{opus}, \textit{tico19} \cite{opus} , \textit{ted20} \cite{reimers-gurevych-2020-making}, \textit{qed} \cite{qed, opus}, and \textit{os} \cite{opus}.\footnote{\url{https://www.opensubtitles.org/}} We group these data sources together and call them clean data.

\paragraph{Back-translation data}
Back-translation is a simple yet effective data augmentation technique that could benefit the training of sentence encoders. To generate back-translation data, we need a translation model and monolingual data. We retrieve monolingual data from web data provided by CommonCrawl\footnote{In wet format, \url{https://commoncrawl.org}} and ParaCrawl \cite{paracrawl}, following the pre-processing pipeline (language ID and heuristic filtering) from NLLB \cite{mmt200}. Then we use the 1.3B-parameter NLLB200-Dense\footnote{Open-sourced at: \newline\url{https://github.com/facebookresearch/fairseq/tree/nllb}} model \cite{mmt200} to translate low resource language's monolingual data that we collected from web. We choose NLLB200 because it achieves state-of-the-art translation performance so far for most low-resource languages. Once we generate synthetic data, we can use them as a parallel corpus to distill sentence representation. In practice, we cap the amount of back-translation data at 3 million lines.

\begin{table}[t]
\centering
\begin{tabular}{cccc}
\toprule
\textbf{ISO} & \textbf{Language} &\textbf{CB[k]} & \textbf{BT[k]}\\
\midrule
khm & Khmer & 536 & 2451 \\
pus & Pashto & 48 & 3000 \\
npi & Nepali & 533 & 3000 \\
sin & Sinhala & 752 & 3000 \\
\bottomrule
\end{tabular}
\caption{\label{wmt-language}
The number of (thousands of) sentences available for different languages in clean bitext (CB column) and back-translation (BT column) datasets. We cap back-translation data to 3 million lines maximum.
}
\end{table}

\subsection{Languages \& Models}
We target four low-resource languages (Khmer, Pashto, Nepali, and Sinhala) in our experiments as shown in Table \ref{wmt-language}. 
For each of these languages, we have trained several models shown below:

\begin{itemize}
    \item LASER \cite{laser}: we directly take the model checkpoint available publicly\footnote{\url{https://github.com/facebookresearch/LASER}} without any further training. 
    \item LASER3 \cite{laser3}: we follow \citet{laser3} to distill an encoder for each language using \textbf{clean data}. The teacher encoder is LASER and the student encoder is a randomly initialized transformer's encoder.
    \item LASER3-BT: same as the LASER3 model except that we distill the encoder using both \textbf{clean data} and \textbf{back-translation data}. Though back-translation is already widely used, our improvement from this model provides another instance where data augmentation from a large pre-trained multilingual translation model is effective.
    \item LASER3-CO: we fine-tune the LASER3 model following \Cref{sec::moco} with \textbf{clean data}.
    \item LASER3-CO-Filter: we fine-tune the LASER3 model following \Cref{sec::laser3-cf} with \textbf{clean data}. 
\end{itemize}

\begin{figure*}[h]
    \centering
    \begin{tabular}{cccc}
    \includegraphics[width=0.25\textwidth]{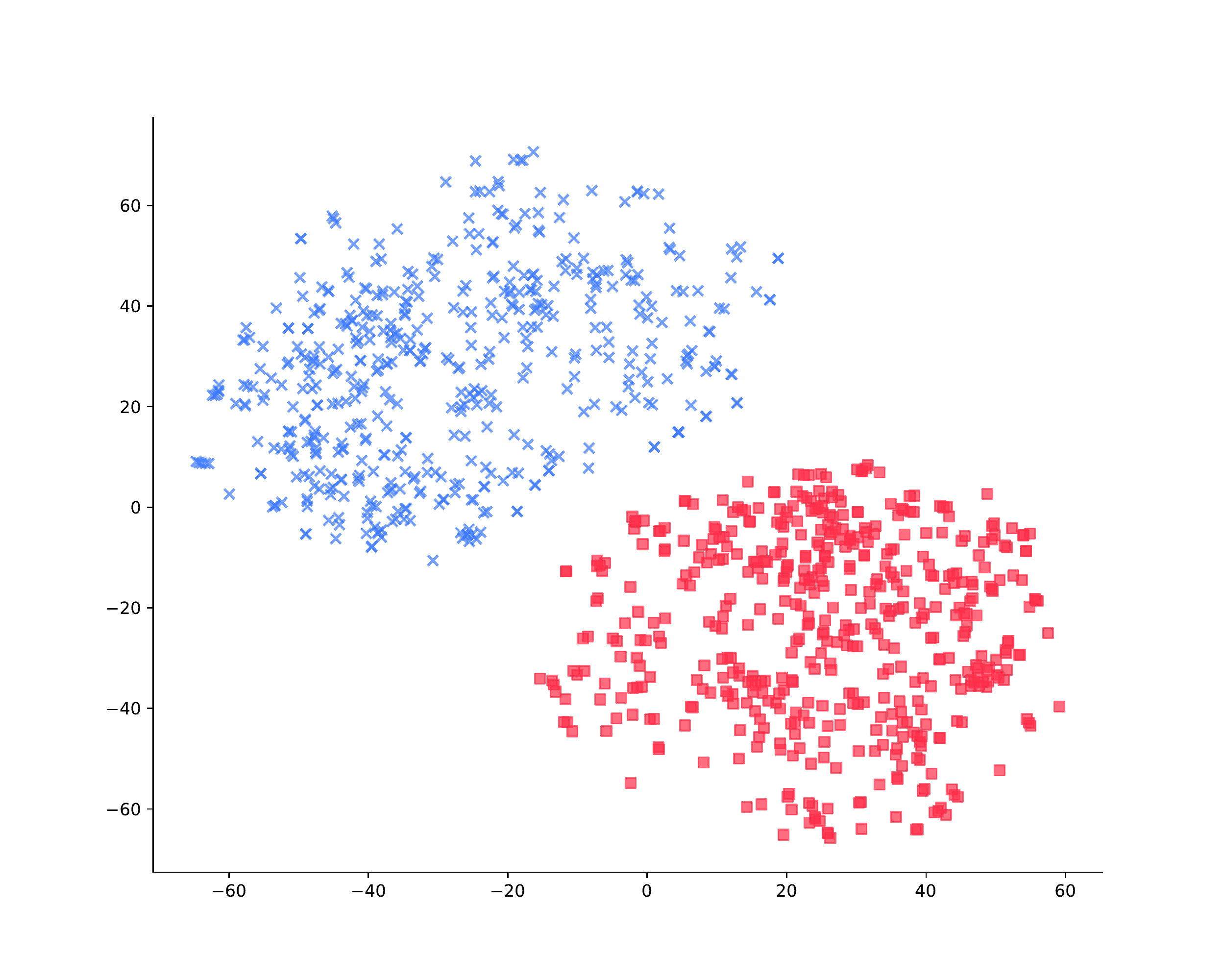} \includegraphics[width=0.25\textwidth]{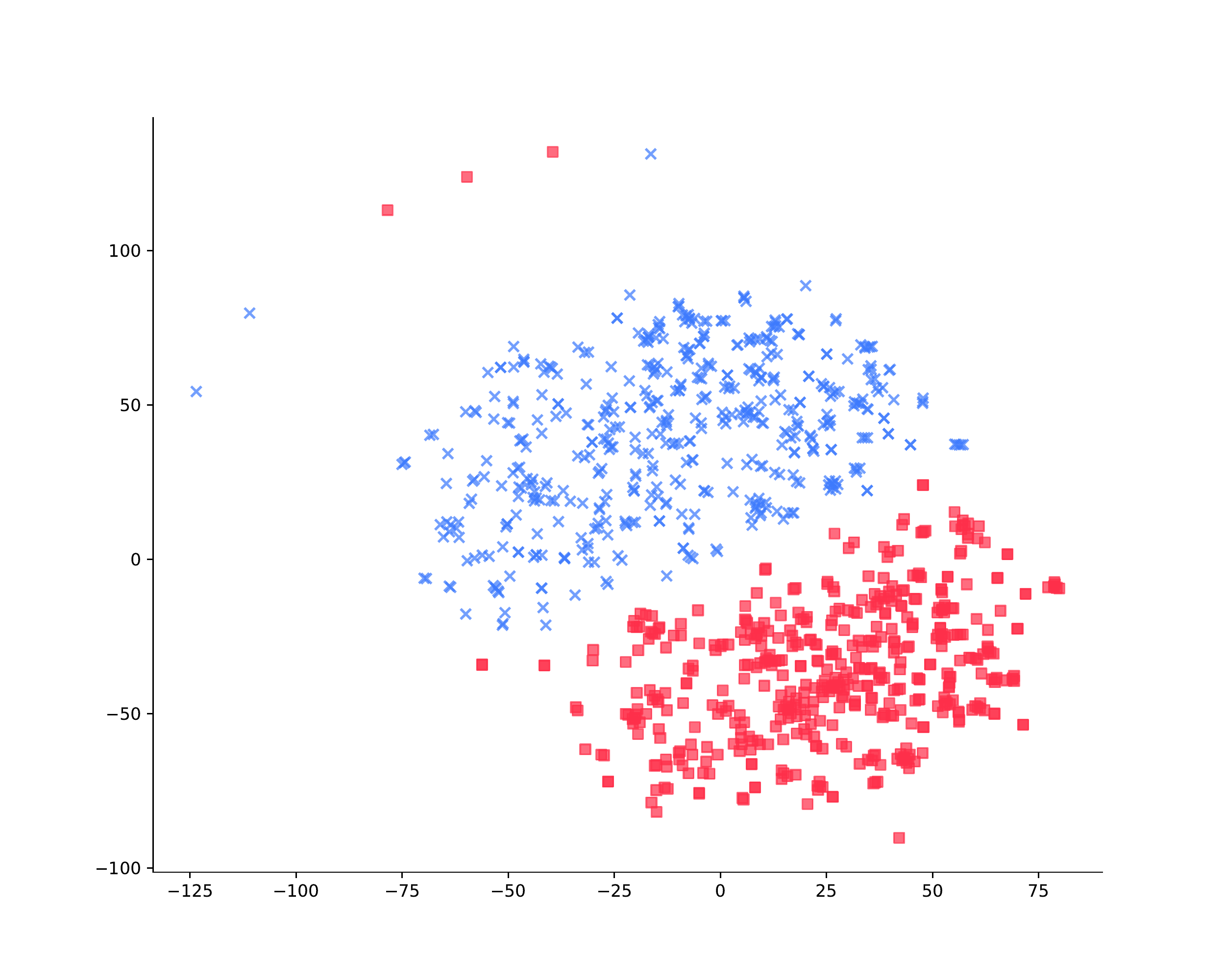} \includegraphics[width=0.25\textwidth]{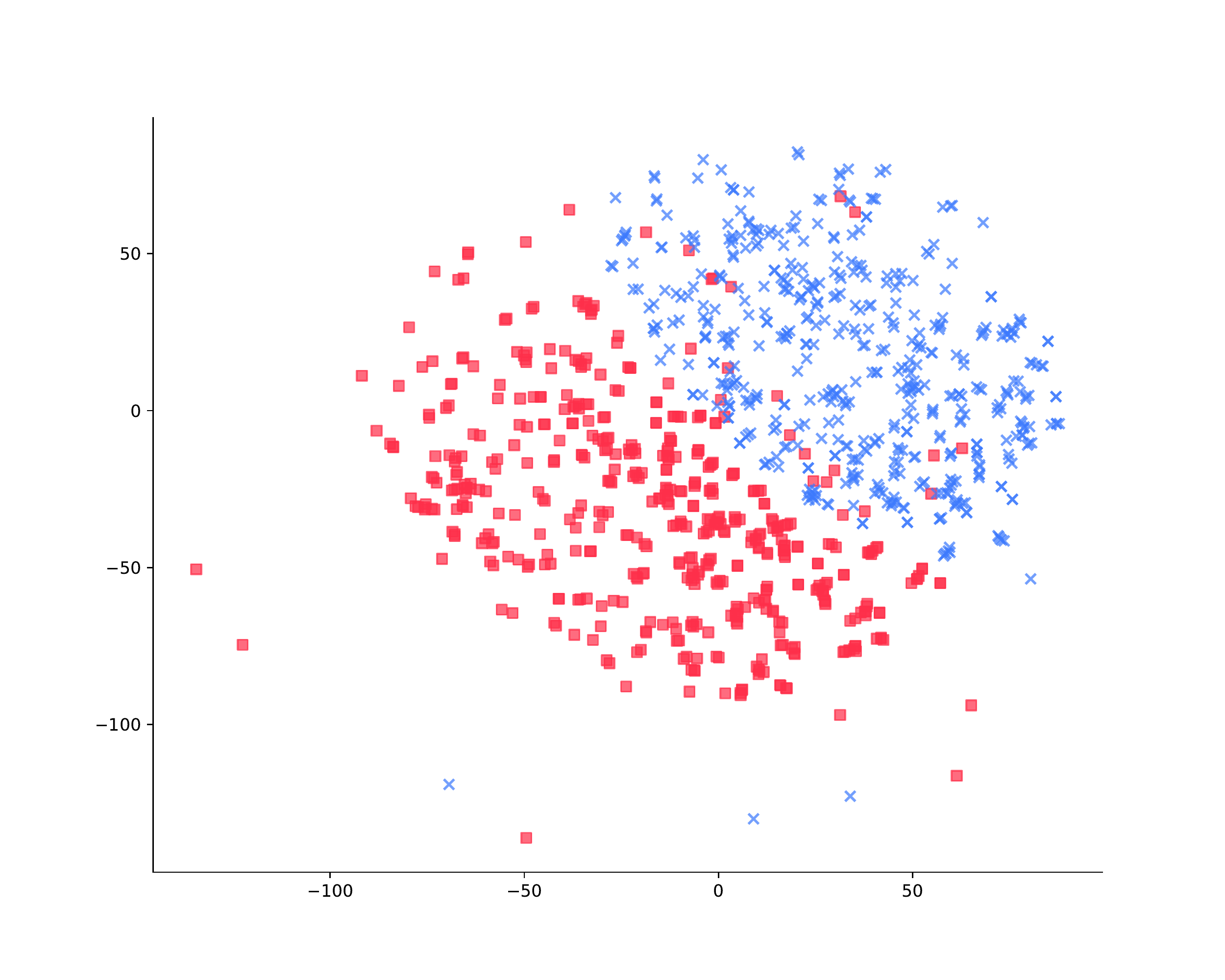} \includegraphics[width=0.25\textwidth]{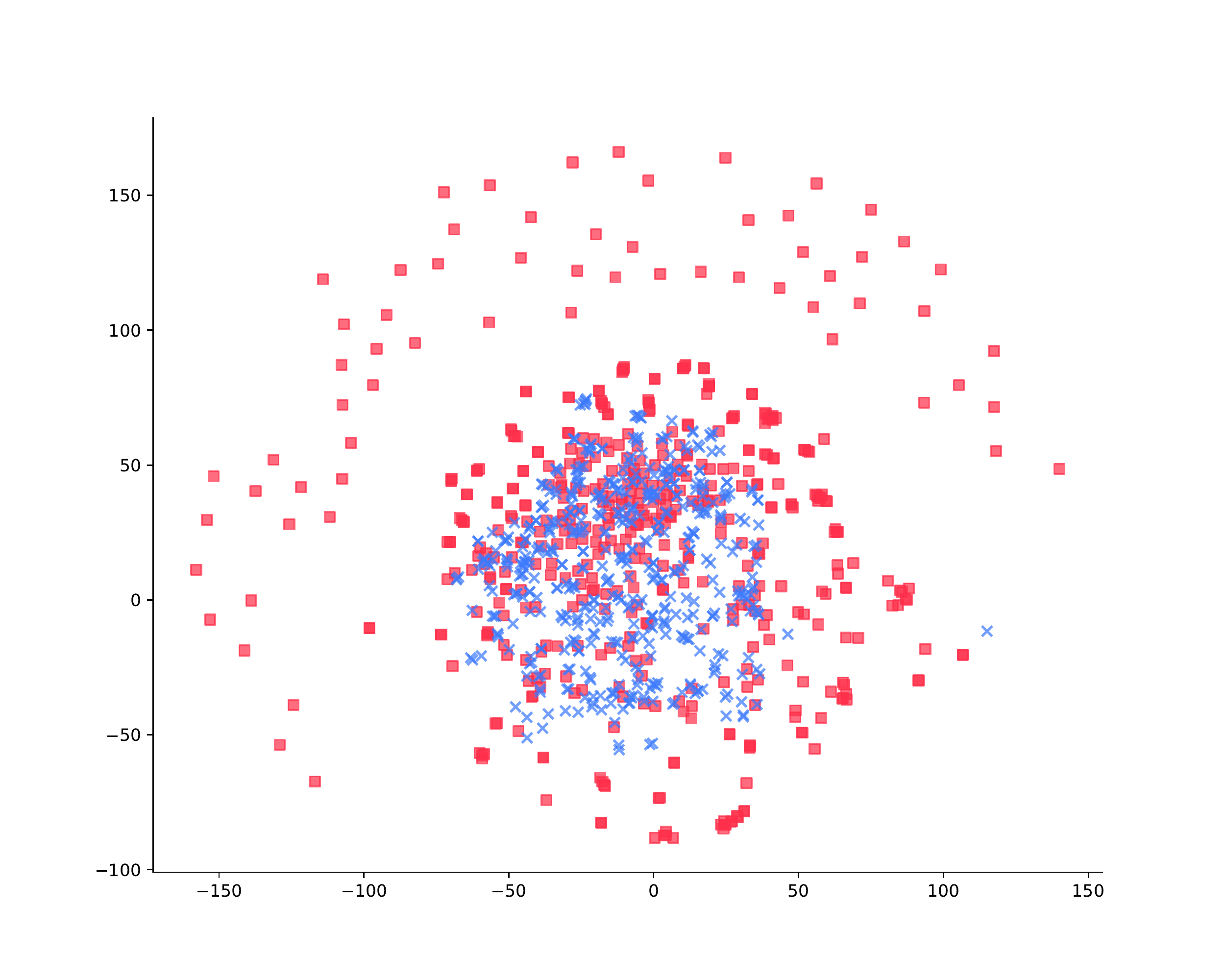} \\
    (a) LASER3\qquad \qquad \quad (b) LASER3-BT \qquad\qquad(c) LASER3-CO\qquad\quad (d) LASER3-CO-Filter\\
    \end{tabular}
    \caption{t-SNE visualizations of embedding space for Sinhala-English. Red points are from Sinhala sentences and blue points are from English sentences.
    \label{fig:analysis-vis}}
\end{figure*}

\subsection{Multilingual Similarity Search}\label{sec::xsim}
Multilingual similarity search evaluates the quality of representation from trained encoders by aligning source and target sentences. We use FLORES \cite{flores} devtest dataset (1012 aligned sentences) for this task because it is crafted with professional human annotation. For every source sentence $x_{i}$, we compute a similarity score called \texttt{xsim} between $x_{i}$ and all target sentences $y_j$. If the target sentence with the highest similarity score is not the $i^{th}$ sentence (namely if $i \neq \text{argmax}_j \texttt{xsim}(x_i, y_j)$), then the model has one incorrect match. We compute the percentage of source sentences that have an incorrect match as \texttt{xsim} error rate. The lower the \texttt{xsim} error rate, the better the model is for aligning sentences of similar meaning.
Our similarity score \texttt{xsim} is the margin-based score \cite{marginscore}, which is widely used for scoring and filtering parallel sentences. It is defined below:
\begin{equation}\label{fnc::xsim}
    \begin{split}
        &\text{xsim}(x,y) = \text{margin}(\cos(x,y), \\
        &\sum_{z \in \text{NN}_k(x)} \frac{\cos(x,z)}{2k} + \sum_{z \in \text{NN}_k(y)} \frac{\cos(y,z)}{2k})
    \end{split}
\end{equation}
where $x, y$ are source and target sentences' representations from the model, $\text{NN}_k(x)$ denotes k nearest neighbors of x in the other language, and there are three margin functions\footnote{We follow \citet{laser3} and \citet{mmt200} to use ratio as the margin.}: $\text{absolute}(a,b) = a$, $\text{distance}(a,b) = a-b$, and $\text{ratio}(a,b) = \frac{a}{b}$. 

We show our result in Table \ref{result:wmt-xsim} and observe that LASER3 achieves a much better result than LASER (1.7\% versus 27.4\%), consistent with \citet{laser3}'s finding that language-specific distillation is more helpful than a language-agnostic encoder. We also show that using data augmentation (either with back-translation or with contrastive learning) further improves the performance to a near-zero error rate. In Figure \ref{fig:analysis-vis}, we provide the t-SNE visualization of embedding spaces trained by different approaches for Sinhala, and we observe that using back-translation or contrastive learning brings aligned sentences closer.

Note that in both of our contrastive-based models (LASER3-CO and LASER3-CO-Filter), we do not include back-translation data. Thus, contrastive distillation achieves tied performance compared to the model helped by a large amount of back-translation data, showing the power of self-supervision in representation learning. Additionally, back-translation and contrastive learning are orthogonal so we could train LASER3-CO/LASER3-CO-Filter with back-translation data. However, in our preliminary study, we do not see improvement from such an approach (more discussion in \Cref{sec::discuss-btcl}) and we use only clean data for contrastive distillation throughout our experiments.

\begin{table}[t]
\centering
\scalebox{0.8}{
\begin{tabular}{c|cccc|c}
\toprule
\textbf{Model} & \textbf{khm} &\textbf{pus} &\textbf{npi} & \textbf{sin} & \textbf{avg}\\
\midrule
LASER & 14.6 & 67.8 & 27.0 & 0.1 & 27.4\\
LASER3 & 0.4 & 0.3 & 5.9 & 0.3 & 1.7\\
LASER3-BT & \textbf{0.1} & \textbf{0.0} & \textbf{0.4} & 0.1 & \textbf{0.2}\\
LASER3-CO & 0.3 & 0.1 & 0.7 & \textbf{0.0} & 0.3\\
LASER3-CO-Filter & \textbf{0.1} & 0.1 & 0.8 & 0.1 & 0.3\\
\bottomrule
\end{tabular}}
\caption{\label{result:wmt-xsim}
\texttt{xsim} error rate (\%) for Khmer, Pashto, Nepali, and Sinhala. Bolded values are the lowest error rate obtained for each language.
}
\end{table}

\subsection{Corpus Filtering}
We pick Khmer, Pashto, Nepali, and Sinhala as our targeted low-resource languages partly because they are used by WMT Corpus Filtering Shared task \cite{wmt19, wmt20}. The shared task release mined corpora (from the Paracrawl\footnote{\url{https://paracrawl.eu/}} project) for Khmer, Pashto, Nepali, and Sinhala. For each language, we first filter the noisy corpus with the language id model from fasttext\footnote{\url{https://fasttext.cc/docs/en/language-identification.html}} \cite{joulin2016bag}. We filter out the aligned sentence pairs whose source sentence is predicted to be English (because we only expect the target sentence to be English). Then we filter the corpus with our sentence encoders where we rely on \texttt{xsim} score to rank sentences and take the top k sentences to train NMT models. In practice, we follow the shared task to use data subset of 1/2/3/5/7 million English tokens, and the optimal subset size is shown in Table \ref{subset}. The highest BLEU score achieved by different systems is shown in Table \ref{result:wmt-mine}.

\begin{table}[t]
\centering
\scalebox{0.75}{
\begin{tabular}{c|cccc}
\toprule
\textbf{Model} & \textbf{khm} &\textbf{pus} &\textbf{npi} & \textbf{sin}\\
\midrule
$\dagger$LASER & 5 & 5 & N/A & N/A \\
$\dagger$WMT Best System & 5 & 5 & 1 &1 \\
LASER3 & 7 &5&1&2 \\
LASER3-BT & 7&5&2&5\\
LASER3-CO & 7&5&2&5\\
LASER3-CO-Filter & 7&7&2&5\\
\bottomrule
\end{tabular}}
\caption{\label{subset}
Optimal subsample size (in millions of tokens) for corpus filtering task. For most cases, 5 or 7 million tokens give the best performance except for Nepali, where only 1 or 2 million tokens seem to be useful.\newline$\dagger$: number taken from findings of WMT corpus filtering shared task \cite{wmt19, wmt20}.
}
\end{table}

From the results, we observe that LASER3-based systems have similar BLEU scores for Khmer. One possible explanation is that LASER3 distillation already gives good representation so data augmentation does not make a noticeable improvement. On average, we find that data-augmented systems (LASER3-BT, LASER3-CO, LASER3-CO-Filter) greatly improve the BLEU score compared to previous works like LASER, LASER3, and best systems in the WMT Corpus Filtering shared task. Note that some systems from the shared task make use of large pre-trained models and even an ensemble of large models, so it is a strong baseline and we are able to outperform it by an average of 0.75 BLEU score, achieving the state-of-the-art corpus filtering performance using contrastive distillation on a small amount of clean data. The corpus filtering result is also consistent with the multilingual similarity search task, where encoders with lower \texttt{xsim} error rate perform better.

\begin{table}[t]
\centering
\scalebox{0.75}{
\begin{tabular}{c|cccc|c}
\toprule
\textbf{Model} & \textbf{khm} &\textbf{pus} &\textbf{npi} & \textbf{sin} & \textbf{avg}\\
\midrule
$\dagger$LASER & 7.1 & 9.7 & N/A & N/A \\
$\dagger$WMT Best System $^*$ & 8.9 & 10.9 & \textbf{6.9} & 6.5 & 8.30 \\
LASER3 & \textbf{10.1} & 11.7 & 5.1 & 5.8 & 8.18 \\
LASER3-BT & \textbf{10.1} & 11.8 & 6.2 & \textbf{7.7} & 8.95\\
LASER3-CO & 9.9 & \textbf{12.3} & 6 & 7.3 & 8.88\\
LASER3-CO-Filter & \textbf{10.1} & 12.2 & 6.2 & \textbf{7.7} & \textbf{9.05}\\
\bottomrule
\end{tabular}}
\caption{\label{result:wmt-mine}
BLEU score for Khmer, Pashto, Nepali, and Sinhala. Bolded values are the highest BLEU scores obtained for each language. $\ast$: The highest number achieved by WMT19 \& 20 corpus filtering task participants. $\dagger$: results taken from findings of WMT19 \& 20 \cite{wmt19, wmt20}.
}
\end{table}

\section{Experiment on Extremely Low-resource Languages}
\subsection{Languages \& Dataset}
In \Cref{sec::low-resource}, we experiment on low-resource languages \textbf{individually} and show the improvement from contrastive distillation. There are many languages that have even less available data and are under-represented in the current literature. In our work, we focus on four extremely low-resource African languages that are extremely low-resource: Chokwe, Kamba, Kimbundu, and Umbundu. The amount of available clean and back-translation data is shown in Table \ref{data-size} (with the same collection process described in \Cref{sec::dataset}). These four languages are from the Benue-Congo language family which has 25 languages in total. The details of their names and available data size are shown in Table \ref{data-size-all} (in the appendix). Under such extremely low-resource settings, we show that contrastive distillation still produces state-of-the-art encoders for the multilingual similarity search task. We believe our encoders can be easily adapted to other information extraction tasks such as bitext mining through the pipeline described in CCMatrix \cite{ccmatrix}, and we leave this to future work.

\begin{table}[t]
\centering
\scalebox{0.93}{
\begin{tabular}{cccc}
\toprule
\textbf{ISO} & \textbf{Language} & \textbf{CB[k]} &\textbf{BT[k]}\\
\midrule
cjk & Chokwe & 40 & 166 \\
kam & Kamba&58 & 74 \\
kmb &Kimbundu& 101 & 140\\
umb &Umbundu& 234 & 366\\
\bottomrule
\end{tabular}}
\caption{\label{data-size}
the number of (thousands of) sentences available for different African languages in clean bitext (CB column) and back-translation (BT column) datasets. We cap back-translation data to 3 million lines maximum.
}
\end{table}

\subsection{Multilingual Similarity Search}\label{sec::african_xsim}
We evaluate the performances of difference encoders on the multilingual similarity search task following the same procedure in \Cref{sec::xsim} and show \texttt{xsim} error rate in Table \ref{result:african-xsim}. For the language-agnostic model baseline, we replace LASER with LaBSE \cite{labse} because its pre-training data contains African languages (though not directly on the four languages we target) and achieves better performance than LASER. However, this language-agnostic model is still easily outperformed by LASER3-based models since the distillation of language-specific data is helpful.

For LASER3 and LASER3-BT, we leverage all data from the Benue-Congo family to distill the encoder. We are able to reproduce results from LASER3 \cite{laser3} and our back-translation baseline further improves the average \texttt{xsim} error rate to 6.7\%. For contrastive distillation models, we fine-tune LASER3-BT \textbf{individually}\footnote{Consistent with our experiment in \Cref{sec::xsim}, using only clean data works the best for contrastive distillation.} for each target language following \Cref{sec::approach}. Not like previous experiments with Khmer, Pashto, etc., we fine-tuned LASER3-BT instead of LASER3 as our preliminary study showed better performance with LASER3-BT. Therefore, different languages could have different optimal settings for data augmentation (more discussion in \Cref{sec::discuss-btcl}), depending on the amount and quality of available data. Nevertheless, we show that our contrastive distillation framework greatly improves upon previous encoders, achieving state-of-the-art \texttt{xsim} error rates for all four languages we targeted.

\begin{figure*}[t]
    \centering
    \begin{tabular}{cc}
    \includegraphics[width=0.5\textwidth]{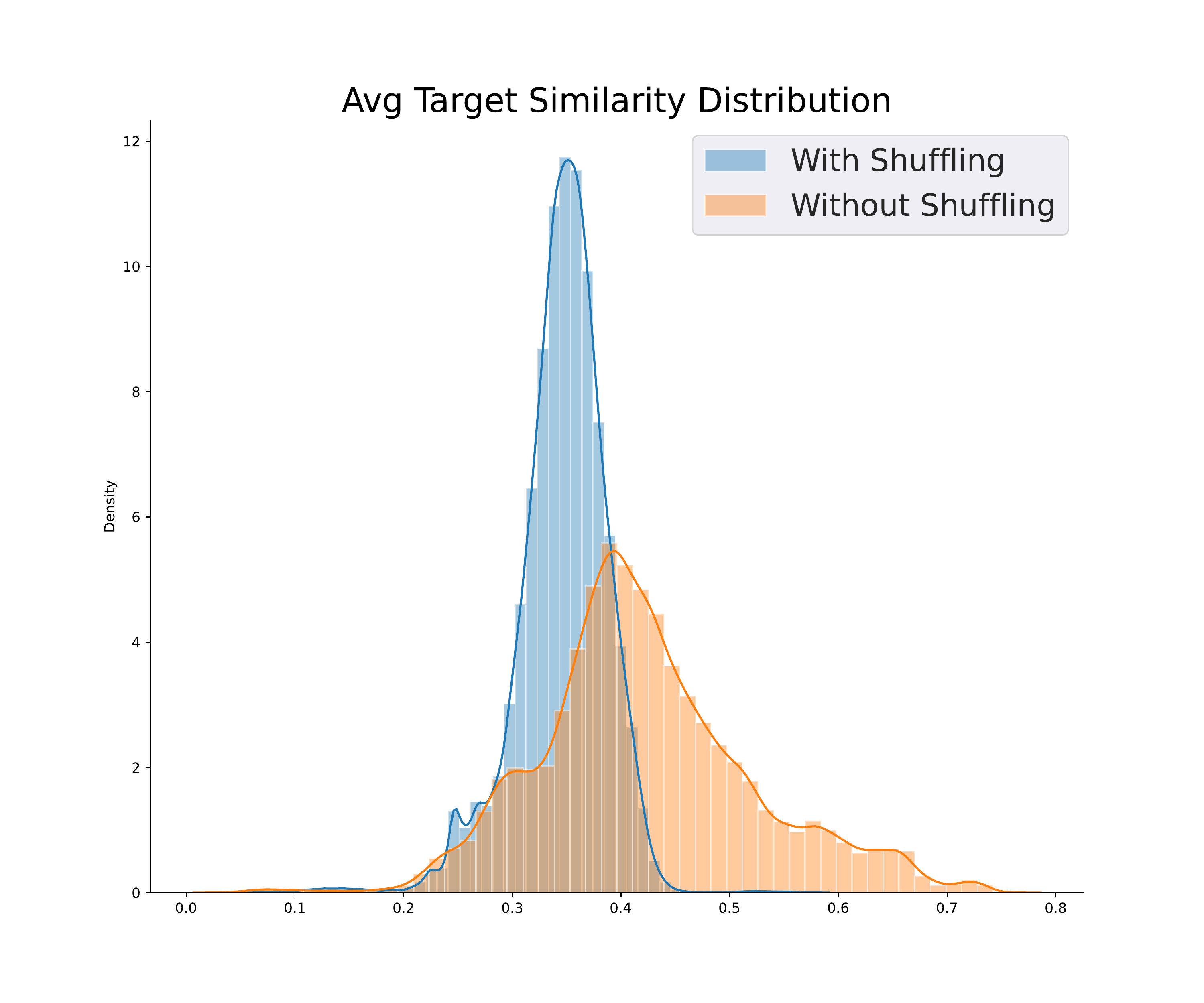} &
    \includegraphics[width=0.5\textwidth]{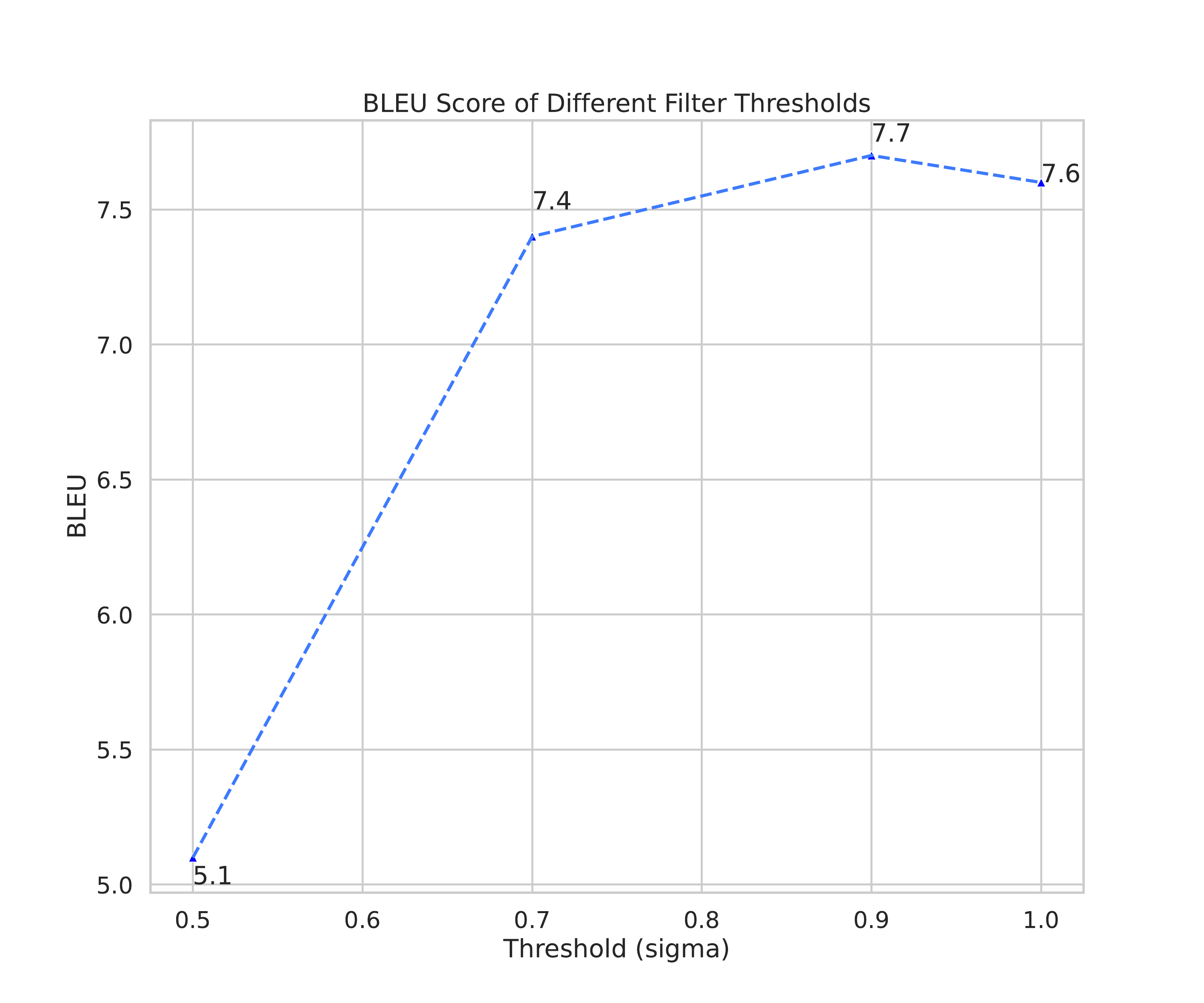}\\
    \end{tabular}
    \caption{(left) Distribution plot of the average target similarity (of Sinhala's clean data) with/without shuffling. The x-axis is the value of average target similarity and the y-axis is the density of a particular value. (right) LASER3-CO-Filter's performance for Sinhala w.r.t varying pre-filter threshold $\sigma$. A low threshold filters out all hard negatives and harms training. A high threshold keeps extremely hard negatives which also undermines training. 
    \label{fig:analysis-sim}}
\end{figure*}

\begin{table}[t]
\centering
\scalebox{0.8}{
\begin{tabular}{c|cccc|c}
\toprule
\textbf{Model} & \textbf{cjk} &\textbf{kam} &\textbf{kmb} & \textbf{umb} & \textbf{avg}\\
\midrule
LaBSE$^*$ & 34.4 & 27.4 & 35.0 & 37.0 & 33.5\\
LASER3$^*$ & 16.4 & 15.3 & 7.5 & 15.6 & 13.7\\
LASER3-BT & 14.6 & 3.5 & 4.8 & 7.0 & 6.7\\
LASER3-BT-CO & \textbf{9.1} & \textbf{1.9} & 1.3 & \textbf{4.6} & \textbf{4.2}\\
LASER3-BT-CO-Filter & 11.3 & 2.2 & \textbf{1.2} & 7.1 & 5.5\\
\bottomrule
\end{tabular}}
\caption{\label{result:african-xsim}
\texttt{xsim} error rate (\%) for Chokwe, Kamba, Kimbundu, and Umbundu. Bolded values are the lowest error rate obtained for each language. $\ast$: result taken from \citet{laser3}.
}
\end{table}

\section{Analysis}\label{sec::analysis}
In this section, we focus on contrastive distillation methods described in \cref{sec::approach} and provide an analysis of the effects of different hyper-parameters.

\subsection{Effect of queue size}
For all the previous experiments, we use batch size 32 and queue size 4096 to be consistent. In our preliminary work, we tried N=512/1024/2048/4096 for Sinhala and we do not observe a noticeable advantage for using a larger queue size. Still, we use the largest queue size that we could afford,\footnote{We trained our models with NVIDIA A100 GPU.} 4096, throughout experiments since a larger queue size won't hurt performance. To show that memory-based contrastive learning is effective, we compare it with in-batch fine-tuning and show the result in Table \ref{result:wmt-inbatch}. We see that memory-based contrastive learning gives us an average of 0.4 BLEU improvement across four languages. 

\begin{table}[h]
\centering
\scalebox{0.8}{
\begin{tabular}{c|cccc|c}
\toprule
\textbf{Model} & \textbf{khm} &\textbf{pus} &\textbf{npi} & \textbf{sin} & \textbf{avg}\\
\midrule
LASER3 & \textbf{10.1} & 11.7 & 5.1 & 5.8 & 8.18 \\
LASER3-CO & 9.9 & \textbf{12.3} & \textbf{6} & \textbf{7.3} & \textbf{8.88}\\
LASER3-InBatch & 9.8 & 11.4 & 5.8 & 6.9 & 8.48 \\
\bottomrule
\end{tabular}}
\caption{\label{result:wmt-inbatch}
BLEU score for Khmer, Pashto, Nepali, and Sinhala. LASER3-InBatch and LASER3-CO have the same setup except that LASER3-CO uses a queue of 4096 samples to compute contrastive loss and LASER3-InBatch only uses samples from every batch to compute contrastive loss.
}
\end{table}

\subsection{Effect of shuffling}
In our LASER3-CO-Filter approach, we disable shuffling so that the queue contains sentences of similar length and we argue that it is easier to find hard negatives this way. To quantify this fact, we compute the average target similarity for all sentences and plot the distribution in Figure \ref{fig:analysis-sim} (left). The average target similarity for a target sentence x is $$\text{sim}(x, \text{queue}) = \frac{1}{N} \cos (\theta_t(x), \theta_t (\text{queue}_i))$$
where $\theta_t$ is the teacher encoder, N is queue size, and $\text{queue}_i$ is the $i^{\text{th}}$ sample in the queue. As shown in Figure \ref{fig:analysis-sim} (left), without shuffling, more than half samples have $>0.4$ average target similarity. A higher average target similarity indicates that target sentences are more semantically related to the samples in the queue and therefore form harder negatives.

\subsection{Effect of pre-filter threshold}
For the LASER3-CO-Filter model, we employ a pre-filtering mechanism that prevents extremely hard examples from contributing to the loss. For our experiments, we use filter threshold $\sigma=0.9$. In this section, we vary $\sigma$ and evaluate the performance for Sinhala (we did not repeat this experiment for other languages but expect similar results). The result is shown in Figure \ref{fig:analysis-sim} (right), which is a line plot based on our result for $\sigma=0.5/0.7/0.9/1$. We can see that $\sigma=0.9$ is the peak and the result corresponds to our intuition. When $\sigma$ is too high (e.g. $\sigma=1$), extremely hard negatives (or even repetitions) are not filtered out and their contribution to contrastive loss makes training unstable, resulting in poor performance for mining. When $\sigma$ is too low (e.g. $\sigma=0.5$), too many samples in the queue are filtered out and only easy negatives (sentences drastically different from the target sentence) are left in the queue. In this case, the model does not learn meaningful representations from negative samples, resulting in poor performance as well.

\section{Related Work}
\paragraph{Multilingual Sentence Representation} Various multilingual sentence representations have been proposed including multilingual BERT \cite{bert}, XLM \cite{xlm}, XLM-R \cite{xlmr}. These models are trained with a large amount of data across dozens or hundreds of languages. They show great improvement when fine-tuned on downstream tasks. However, they are trained only with token-level objectives like masked language modeling (MLM) or translation language modeling (TLM) and their sentence representations cannot be directly used for semantic similarity tasks. To address this problem, different methods make use of pooled sentence embedding to train better encoders \cite{laser, sbert, labse, laser3}. 


\paragraph{Back-translation} Back-translation is widely used in machine translation as a data augmentation technique. \citet{backtranslation} shows that synthetic data created by translating monolingual data greatly improve English-German and Turkish-English translation models. \citet{btscale} further investigate different decoding strategies and show that sampling works better than greedy decode or beam search for back-translation.
Since back-translation's quality depends on the translation model, \citet{iterativebt} proposes iterative back-translation where translation models are further improved by the synthetic data they produce.

\paragraph{Contrastive Learning} Contrastive learning is a form of self-supervised learning that encourages similar inputs to have close representations and dissimilar inputs to have different representations. \citet{simclr} proposes SimCLR that uses contrastive learning to improve visual representation. \citet{supcon} integrates supervised learning into contrastive learning and \citet{tian2019crd} uses contrastive distillation for image representation learning. Since finding high-quality contrastive samples is essential for contrastive learning, \citet{memorycon} uses a memory bank to store contrastive samples while \citet{moco} uses a queue with momentum update to encode contrastive samples on the fly.  Though previously-mentioned works are evaluated on computer vision tasks, there is growing interest in using contrastive learning for natural language processing and various approaches have been proposed to train better sentence representation \cite{giorgi-etal-2021-declutr, Kim2021SelfGuidedCL, peng-etal-2020-huaweis, simcse, bicon}.

\section{Conclusion}
In this paper, we integrate contrastive learning into multilingual sentence representation distillation. Through self-supervision from negatives, especially hard negatives, we obtain better encoders. Compared to other effective data augmentation techniques like back-translation, our contrastive distillation method achieves tied or better performance with less computation and data. Our experiments on multilingual similarity search and corpus filtering tasks show consistent improvement from contrastive learning and achieve state-of-the-art performance for various low-resource languages.

\section*{Limitations}
This research focuses on improving cross-lingual encoders to perform quality estimation of parallel data for low-resource machine translation. However, cross-lingual representation could be helpful in many other tasks as summarized in the XTREME benchmark \cite{xtreme1,xtreme2} which we have not experimented with. Another limitation is that we disabled shuffling to store hard negative samples in the queue. However, without shuffling, it might affect the stability of the training process (although we do not observe it in our experiments). Therefore, in future work, we will continue exploring other cost-efficient approaches to retrieve hard negatives. 

\section*{Acknowledgements}
We thank anonymous reviewers for their valuable comments. We also thank Alex Guo, Huda Khayrallah, Kaushik Ram Sadagopan, Simeng Sun, Guillaume Wenzek, and Haoran Xu for their helpful suggestions.

\bibliography{anthology,custom}
\bibliographystyle{acl_natbib}

\clearpage

\appendix

\section{PyTorch-Style Pseudocode}\label{sec::pseudocode}
In Algorithm \ref{algo:filter}, we provide the pythonic style pseudocode for our LASER3-CO-Filter model. Note that the pre-filtering mechanism needs special treatment when training with a minibatch. we compute the set of negative samples $$S^j = \{i: \cos(k_{+}^j, k_i) < \sigma\}$$ for each target sentence $j$ in our batch where $k_{+}^j$ is the teacher encoding of $j^{th}$ positive sample. Then we find the minimum number of negatives in the batch $M = \text{min}_j |S^j|$. For any target sentence $j$ that has $|S^j| > M$, we randomly remove $|S^j| - M$ samples so that it ends up with $M$ samples. In this way, we ensure that every target sentence has the same number of legitimate negative samples after the pre-filtering mechanism and we can batch them to compute the InfoNCE loss. In the pseudocode, we abstract the implementation of pre-filtering using \textbf{filter} function.

\section{Corpus Filtering Threshold}\label{sec::filter_threshold}
In Table \ref{subset}, we show the optimal size of filtered corpus to train NMT models for four languages. For most cases, 5 or 7 million tokens give the best performance (except for Nepali, where only 1 or 2 million tokens seem to be useful).

\section{Combine Back-translation and Contrastive Learning?}\label{sec::discuss-btcl}
Back-translation and contrastive learning are two helpful ways for representation learning in low-resource settings. These two methods can be applied together (as we showed in our experiments on African languages) because back-translation prepares data before training while our contrastive distillation method is a fine-tuning strategy. Through our preliminary study, we made the following observations:
\begin{itemize}
    \item Contrastive fine-tuning works best on a single language's clean data. This is true for both low-resource languages (Khmer, Pashto, Nepali, and Sinhala) and extremely low-resource languages ( Chokwe, Kamba, Kimbundu, and Umbundu). We find that using back-translation data for contrastive objectives does not make any noticeable improvement and this is probably because synthetic data generated by the pre-trained NMT model does not produce too many hard negatives. In the case of the African languages, we tried distillation with the whole family's clean data and it is also less effective than distillation on a single language. This is probably because the data from multiple languages interfere with each other during contrastive distillation. 
    \item For low-resource languages like Sinhala, Pashto, etc., we find no noticeable difference between contrastive distillation with LASER3 and LASER3-BT. On the contrary, for extremely low-resource languages, we find LASER3-BT-CO performs better than LASER3-CO, and this is why we used LASER3-BT-CO in \Cref{sec::african_xsim}  but LASER3-CO in \Cref{sec::xsim}. We believe this is due to the quality and the amount of available clean data. In the extremely low-resource case, even the limited amount of clean data contains noise and we see LASER3-BT greatly outperform LASER3. Therefore, the distillation with pre-trained LASER3-BT result in a better model (LASER3-BT-CO or LASER3-BT-CO-Filter). However, in some low-resource settings (like our experiments for Sinhala, Pashto, Khmer, and Nepali), clean data itself could allow the contrastive method to learn good representations, therefore avoiding the need for back-translation data.
\end{itemize}
With our observation above, how should we select methods when new languages come in? We believe it mainly depends on two factors: 1)~How large and how clean is the existing clean corpus 2)~How is the performance of the existing pre-trained NMT model?

The first factor affects the effectiveness of contrastive learning and the second factor affects the quality of back-translation data. In fact, in our preliminary study, we do not find back-translation from a statistical machine translation system useful for African languages, because it does generate good enough synthetic data. With the advance of multilingual machine translation (MMT) models, we are able to leverage a large pre-trained model and show its effectiveness. Thus, whether to add back-translation or not mainly depends on the translation model's quality and computation budget.

\clearpage
\begin{table*}[t]
\centering
\begin{tabular}{cc|cc||cc|cc}
\toprule
\textbf{ISO} & \textbf{lang} & \textbf{CB[k]} &\textbf{BT[k]} &\textbf{ISO}&\textbf{lang} & \textbf{CB[k]} &\textbf{BT[k]}\\
\midrule
khm & Khmer & 536 & 2451 & pus &Pashto & 48 & 3000\\
npi & Nepali & 533 & 3000 & sin & Sinhala & 752 & 3000\\
cjk & Chokwe & 40 & 166 & kam & Kamba & 58 & 74 \\
kmb & Kimbundu & 101 & 140 & umb & Umbundu & 234 & 366\\
bem & Bemba & 700 & 828 & ibo & Igbo & 912 & 3000 \\
kik & Kikuyu & 119 & 95 & kin & Kinyarwanda & 517 & 3000 \\
kon & Kongo & 227 & 197 & lin & Lingala & 981 & 848 \\
lua & Luba-Kasai & 321 & 492 & lug &  Luganda & 340 & 1654 \\
nso & Northen Sotho & 625 & 902 & nya & Chewa;Nyanja & 867 & 3000\\
run & Rundi & 664 & 1910 & sna & Shona & 789 & 3000 \\
sot & Sotho & 1510 & 3000 & ssw & Swati &116 & 432 \\
swh & Swahili &2778 & 0 & tsn & Tswana & 1748 & 1680 \\
tso & Tsonga & 852 & 1423 & tum & Tumbuka & 264 & 675 \\
xho & Xhosa & 1739 & 3000 & yor & Yoruba & 517 & 3000 \\
zul & Zulu & 2383 & 3000 &&&& \\
\bottomrule
\end{tabular}
\caption{\label{data-size-all}
number of (thousands of) sentences available for different languages in clean bitext (CB column) and back-translation (BT column) dataset.
}
\end{table*}

\begin{table*}[h]
\centering
\begin{tabular}{c|c|c|c|c|c|c|c|c|c|c|c}
\toprule
\textbf{ISO} & \textbf{bible} &\textbf{gv} &\textbf{jw300} &\textbf{qed} & \textbf{tatoeba} &\textbf{ted20} & \textbf{tico19} & \textbf{wikimedia} & \textbf{os} \\
\midrule
khm & \checkmark & \checkmark &\checkmark & \checkmark & \checkmark & \checkmark & \checkmark&\checkmark &\\
pus & & & & \checkmark& \checkmark&\checkmark&\checkmark &\checkmark &\\
npi & \checkmark & \checkmark &\checkmark &\checkmark&\checkmark &\checkmark & &\checkmark &\\
sin & \checkmark & &\checkmark &\checkmark &\checkmark &\checkmark &&\checkmark &\checkmark \\
cjk & & & \checkmark &&&&&&\\
kam & & & \checkmark & &\checkmark &&&&\\
kmb & & &\checkmark &&&&&&\\
umb & & & \checkmark &\checkmark &\checkmark&&&&\\
\bottomrule
\bottomrule
\end{tabular}
\caption{\label{data-source}
Source of clean bitext data for different languages.
}
\end{table*}

\end{document}